\pdfoutput=1

\documentclass[11pt]{article}

\usepackage[preprint]{acl}

\usepackage{times}
\usepackage{latexsym}
\usepackage{amsmath,bm}
\usepackage{graphicx}
\usepackage{booktabs}
\usepackage{caption}
\usepackage{subcaption}
\usepackage{bbm}
\usepackage{anyfontsize}
\usepackage{multirow}

\usepackage{enumitem}

\usepackage[utf8]{inputenc}
\usepackage{lmodern}

\DeclareMathOperator*{\argmax}{arg\,max}

\usepackage[T1]{fontenc}

\usepackage{amssymb}

\usepackage{microtype}

\usepackage{inconsolata}

\title{Muting Whisper: A Universal Acoustic Adversarial Attack on Speech Foundation Models}

\author{Vyas Raina$^*$ \quad Rao Ma$^*$ \quad Charles McGhee \quad Kate Knill \quad Mark Gales \\
}

\begin{document}
\maketitle
\begin{abstract}

Recent developments in large speech foundation models like Whisper have led to their widespread use in many automatic speech recognition (ASR) applications. These systems incorporate `special tokens' in their vocabulary, such as $\texttt{<|endoftext|>}$, to guide their language generation process. However, we demonstrate that these tokens can be exploited by adversarial attacks to manipulate the model's behavior. We propose a simple yet effective method to learn a universal acoustic realization of Whisper's $\texttt{<|endoftext|>}$ token, which, when prepended to any speech signal, encourages the model to ignore the speech and only transcribe the special token, effectively `muting' the model. Our experiments demonstrate that the same, universal 0.64-second adversarial audio segment can successfully mute a target Whisper ASR model for over 97\% of speech samples. Moreover, we find that this universal adversarial audio segment often transfers to new datasets and tasks. Overall this work demonstrates the vulnerability of Whisper models to `muting' adversarial attacks, where such attacks can pose both risks and potential benefits in real-world settings: for example the attack can be used to bypass speech moderation systems, or conversely the attack can also be used to protect private speech data.~\footnote{The code is available at: \url{https://github.com/rainavyas/prepend_acoustic_attack}.}

\end{abstract}

\section{Introduction}

\begin{figure*}[htb!]
    \centering
    \includegraphics[width=\linewidth]{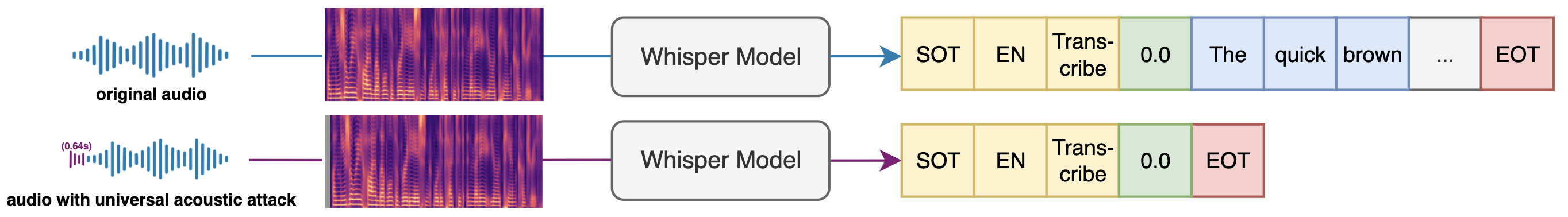}
    \caption{Universal adversarial audio segment when prepended to any speech signal \textit{mutes} Whisper, such that an empty transcription is generated. The $\texttt{<endoftext>}$ token (\textit{EOT}) is a special token in the Whisper vocabulary used to indicate the end of the generated transcription.}
    \label{fig:intro}
\end{figure*}

The development of large foundation models has led to rapid advancements in audio processing, where for example some of the most popular models are of the Whisper family~\citep{radford2022robust}. To guide the generation of natural language, foundation models typically make use of `special' tokens in their vocabulary that do not exist as real text or real acoustic events. As an example, most auto-regressive foundation models will have some form of a $\texttt{<start>}$ token and an $\texttt{<end>}$ token to indicate when to begin generating the output sequence and when to stop. However, we demonstrate that despite their need, these `special' tokens can be exploited by adversaries to make foundational models behave in undesired manners. Specifically, we show that the $\texttt{<endoftext>}$ special token can be exploited by adversaries to prevent an Automatic Speech Recognition (ASR) model, such as Whisper, from transcribing the source audio, i.e., `muting' the model.

Our proposed acoustic adversarial attack method is designed to `mute' Whisper, by learning an extremely short (0.64-second) adversarial acoustic realization of the $\texttt{<endoftext>}$ special token (used by Whisper), where the learnt adversarial audio segment can be prepended to the target speech signal. Furthermore, our proposed method gives a \textit{universal} adversarial audio segment, which allows the \textit{same} 0.64-second adversarial audio segment to be prepended to any speech signal, and conceal its contents from the ASR system, as depicted in Figure \ref{fig:intro}.

Our experiments, conducted across eight different Whisper ASR models, demonstrate that the same universal 0.64-second adversarial audio segment can successfully `mute' Whisper models for more than 97\% of unseen speech samples. We further find that there is a surprising level of transferability of this universal adversarial audio segment to different speech domains (we consider four diverse datasets) and can even transfer to different tasks - the adversarial audio segment can `mute' Whisper when used for speech translation as well as transcription. Muting Whisper has significant implications in high stakes settings. Automatic speech recognition (ASR) systems play a crucial role in detecting and moderating harmful content such as hate speech~\citep{macavaney2019hate} in audio or video recordings~\citep{wu2020detection}. Muting Whisper poses a risk of circumventing this moderation process. Adversaries could exploit this vulnerability to release harmful content to the public audience without detection. Nevertheless, muting Whisper also has potential positive implications for speech privacy protection~\citep{10037237}. In contexts where speech recordings are transmitted over a network, malicious actors may attempt to extract private data through automated transcription. In such cases, our proposed method of muting Whisper could serve as a form of speech privacy protection, similar to a `jamming' signal. Overall, this work demonstrates the vulnerability of Whisper models to muting adversarial attacks, which can have negative or positive implications.

\section{Related Work}
\paragraph{Audio Attacks (early research).} Initial research~\citep{DBLP:journals/corr/abs-1711-03280, cisse2017houdini} explored gradient-based approaches to perturb the input audio to end-to-end ASR systems (specifically WaveCNN and HMM-DNN architectures) with the objective of increasing the word error rate (WER) of the generated transcriptions. However, ~\citet{yuan2018commandersong, DBLP:journals/corr/abs-1801-01944, DBLP:journals/corr/abs-1805-11852, qin2019imperceptible} offer methods to perform targeted attacks on ASR systems, such as DeepSpeech, HMM-DNN and LSTM-based neural networks, where the aim was to generate a specific output transcription. Other research~\citep{schönherr2018adversarial, Schnherr2018AdversarialAA} modified audio adversarial attack methods to better encourage their imperceptibility. 
\paragraph{Practical Audio Attacks.} \citet{DBLP:journals/corr/abs-1905-03828} demonstrate that they can generate \textit{universal} adversarial perturbations such that the same adversarial audio segment can be superimposed on different speech signals. However, these attack approaches cannot be applied to streaming ASR systems, as they have to be superimposed on the entire speech signal, so \citet{10.1145-3372297.3423348} attempted to address this issue by generating universal adversarial perturbations that do not need to be synchronised with the source speech signal (the carrier audio) when being superimposed. \citet{lu2021exploring} extended the targeted universal adversarial attacks to more recent end-to-end ASR systems including LAS, CTC and RNN-T. Further, a range of other creative approaches have been proposed for generating audio adversarial samples in practical settings: transferability from substitute models~\citep{247642, 9348184, ma2021simulating}; evolutionary attacks~\citep{DBLP:journals/corr/abs-1801-00554, khare2019adversarial, taori2019targeted, du2019sirenattack, Zheng_2021}; utterance-based attacks~\citep{raina_gales_knill_2020}; and featurization attacks~\citep{197215, DBLP:journals/corr/abs-1708-09537, abdullah2019practical}.
\paragraph{Attacks on Whisper.} All of the above-mentioned methods are designed for traditional ASR systems. The recent emergence of a powerful foundation model (Whisper) demands an update to previously developed attack methods. \citet{olivier2023kind} perform an initial investigation into the vulnerabilities of Whisper to audio adversarial attacks, where they show that an adversarial signal can be superimposed on natural speech signals such that Whisper transcribes incorrectly.
\paragraph{Our Contributions.} We extend the research on adversarial attacks for modern ASR systems such as Whisper, by outlining a method to develop a truly practical and effective adversarial attack with a real-world targeted objective. Specifically, this work makes the following contributions:
\begin{itemize}[leftmargin=*]
\itemsep0em
    \item We develop a short (0.64-second) adversarial audio segment that can be \textit{prepended} to a speech signal. Existing research tends to consider superimposing the adversarial audio signal, which is not a practical setting for real-world attacks.
    \item Our adversarial audio segment is \textit{universal}, so the same audio segment can be prepended to any speech signal.
    \item Our attack works for a popular, modern and powerful ASR system: Whisper family of models.
    \item The objective of our attack is specifically to \textit{mute} the Whisper model; a targeted objective not before considered and with real-world implications in privacy and security.
    \item Our universal adversarial acoustic attack segment \textit{transfers} across data domains and even speech processing tasks.
\end{itemize}
\section{Speech Processing: Whisper} \label{sec:whisper}

Continuous-time speech is sampled such that the audio can be represented as a sequence of samples, $\mathbf x=x_{1:N}$. An Automatic Speech Recognition (ASR) system maps this sampled speech/audio signal, $\mathbf x$, to the text, $\mathbf y = y_{1:M}$ uttered in the speech signal - this is the transcription of the audio with $M$ words/tokens. Whisper's encoder-decoder architecture, $\mathcal F(\cdot)$ with parameters $\theta$ auto-regressively predicts a vector representing the probability distribution over the vocabulary of tokens, $\mathcal V$, for the next token $y_m$, with the speech, $\mathbf x = x_{1:N}$ at the encoder input and the previously decoded tokens, $\mathbf y^*_{<m}$ at the decoder input,
\begin{equation} \label{eqn:dist}
    P(y_m = y | \mathbf x, \mathbf y^*_{<m}) = \mathcal F(\mathbf x, \mathbf y^*_{<m};\theta)_y,\hspace{1em}y\in\mathcal V,
\end{equation}
where typically a greedy decoding process selects the most likely token to generate,
\begin{equation}\label{eqn:sel}
    y^*_m = \argmax_{y} P(y_m = y | \mathbf x, \mathbf y^*_{<m}).
\end{equation}
During the decoding process various special tokens are used by the Whisper model to guide the token generation. The first token (input to the decoder) is set as \texttt{<|startoftranscript|>}, followed by a token to indicate the language, for example \texttt{<en>} for English. As the Whisper model is trained to perform two different speech processing tasks (transcription and speech translation), the next token is used to indicate the task, e.g., \texttt{<|transcribe|>} or \texttt{<|translate|>}. Hence we define $\mathbf y^*_0 =$ $\texttt{<|startoftranscript|>}$ $\texttt{<lang tag>}$ $\texttt{<|task tag|>}$~\footnote{Note that for the English-only variant of Whisper models, $y^*_0 =$ $\texttt{<|startoftranscript|>}$}. With this initialization, further tokens are generated auto-regressively from the vocabulary, $\mathcal{V}$ following Equation \ref{eqn:dist} and Equation \ref{eqn:sel}. The auto-regressive decoding ends when the \texttt{<|endoftext|>} special token is predicted.

\section{Universal Prepend Attack} \label{sec:attack}

\subsection{Attack Objective}


In this section we propose a practical and effective approach for an adversary to modify any input speech signal in a manner that results in the Whisper model being muted (transcribing nothing), without the speech audio sounding obviously manipulated to human listeners. The objective of muting Whisper is equivalent to maximizing the probability of the model predicting, $y_1$ as the \texttt{<|endoftext|>} special token. Recall that the decoder is initialized with a sequence of special tokens, $\mathbf y^*_0 =$ $\texttt{<|startoftranscript|>}$ $\texttt{<lang tag>}$ $\texttt{<|task tag|>}$.

\subsection{Prepend Attack}
To perturb a speech signal, $\mathbf x = x_{1:N}$, it is simplest to prepend a short, adversarial audio segment of $T$ frames, $\tilde{\mathbf x} = \tilde{x}_{1:T}$, such that the perturbed speech signal is $\tilde{\mathbf x}\oplus\mathbf x$, where $\oplus$ represents concatenation in the raw audio space. Then, given Whisper's encoder-decoder model in Equation \ref{eqn:dist}, the optimal adversarial audio segment, $\hat{\tilde{\mathbf x}}$, to `mute' Whisper as per the adversarial objective, can be given as finding the adversarial audio segment that maximizes the probability of generating the $\texttt{<|endoftext|>}$ special token (abbreviated to $\texttt{eot}$) as the first transcribed token,
\begin{equation} \label{eqn:prepend}
    \hat{\tilde{\mathbf x}}
    = \argmax_{\tilde{\mathbf x}}
    P(y_1 = \texttt{eot} | \tilde{\mathbf x}\oplus\mathbf x, \mathbf y^*_0). 
\end{equation}

\subsection{Universal Attack}
Learning an adversarial audio segment, $\hat{\tilde{\mathbf x}}$ that can be prepended to a speech signal, $\mathbf x$ to conceal its contents from a Whisper ASR model, cannot be achieved in real-time (as the attack segment has to be prepended before the speech is generated) and requires computational resources. Therefore, it is not practical to learn an individual adversarial audio segment $\hat{\tilde{\mathbf x}}^{(j)}$ to conceal the contents of each different speech signal, $\mathbf x^{(j)}$. Hence, we propose learning a \textit{universal} adversarial audio segment that is agnostic to any speech signal. For a training dataset of $J$ speech samples $\{\mathbf x^{(j)}\}_{j=1}^J$, the universal prepend attack aims to maximise the likelihood of predicting $y_1=\texttt{<|endoftext|>}$ over all training samples,
\begin{equation}\label{eqn:universal}
        \hat{\tilde{\mathbf x}} = \argmax_{\tilde{\mathbf {x}}}\prod_{j=1}^J 
        P(y_1 = \texttt{eot} | \tilde{\mathbf x}\oplus\mathbf x^{(j)}, \mathbf y^*_0).
\end{equation}
As the Whisper encoder-decoder model is fully differentiable, standard gradient-based training approaches can then be used to optimize for the universal adversarial audio segment, $\hat{\tilde{\mathbf x}}$. This universal adversarial audio segment `mutes' Whisper when prepended to any speech signal and is thus effectively an acoustic realization of the \texttt{<|endoftext|>} special token.

\subsection{Imperceptibility}
For a truly practical adversarial attack, it is important for the adversarial audio segment generated to be sufficiently imperceptible such that it is not flagged as suspicious when prepended to natural speech signals. We achieve this imperceptibility in two dimensions. First, we ensure that the adversarial audio segment is extremely short such that there is little time for a human listener to detect the abnormal speech. We specifically limit the number of frames in the adversarial audio segment to $T=10240$, which corresponds to 0.64-seconds of audio for a 16kHz sampling frequency. Next, we limit the `power' of the adversarial audio segment, to ensure the amplitude of the adversarial audio segment is not significant relative to natural speech. To limit the power, we introduce a constraint in the optimization objective of Equation \ref{eqn:universal} that limits the amplitude of the adversarial audio,
\begin{equation} \label{eqn:const}
    ||\hat{\tilde{x}}_{1:T}||_{\infty} \leq \epsilon,
\end{equation}
where $||\cdot||_{\infty}$ represents the l-infinity norm. By default we set $\epsilon=0.02$, as on the log-mel scale this empirically represents audio signals with power lower than typical human speech signals (refer to Figure \ref{fig:spect}). The l-infinity norm constraint is incorporated during gradient-based learning of the adversarial audio segment $\hat{\tilde{\mathbf {x}}}$, by clamping the values at $\epsilon$.~\footnote{Clamping after each gradient update is typical in Projected Gradient Descent~\citep{madry2019deep}.} Note that in practical settings it may be undesirable to have extremely low values for $\epsilon$, as the adversarial audio segment may then be contaminated by low-amplitude background noise.

\section{Muting Attack Evaluation}

\subsection{Attack Performance Evaluation}

For a learnt universal acoustic adversarial segment trained to maximize the probability of the Whisper model generating the $\texttt{<|endoftext|>}$ special token as its first token for any speech signal, as per Equation \ref{eqn:universal}, we can evaluate the performance of the adversarial attack by computing the percentage of unseen test speech signals, $\varnothing$, for which the attack is able to successfully `mute' the Whisper model,
\begin{align}\label{eqn:frac0}
    \varnothing &= \frac{1}{J}\sum_j\mathbbm{1} \{{\tilde y^{*(j)}_1 = \texttt{eot}}\}\times 100\%, \\ \nonumber 
    \tilde y^{*(j)}_1 &= \argmax_{y} P(y_1 = y | \hat{\tilde{\mathbf x}}\oplus\mathbf x^{(j)}, \mathbf y^*_0),
\end{align}
where $\tilde y^*_1=$ $\texttt{<eot>}$ means that the transcribed sequence has 0 words, i.e., a perfectly successful attack. Hence, the larger the value of $\varnothing$, approaching 100\%, the more effective the acoustic adversarial attack. A further useful metric to gauge the extent to which a universal attack is able to `mute' the Whisper model, is the `average sequence length' ($\texttt{asl}$) of the predicted transcription,
\begin{equation}\label{eqn:asl}
    \texttt{asl} = \frac{1}{J}\sum_j\texttt{len}(\tilde{\mathbf{y}}^{*(j)}),
\end{equation}
where $\texttt{len}(\cdot)$ gives the number of words in the transcribed sequence. The lower the value of $\texttt{asl}$, the more effective the adversarial attack.

\subsection{Adversarial Sensitivity Analysis} \label{sec:saliency}
Beyond simply measuring the success of the acoustic adversarial attack in `muting' an ASR system, it is meaningful to analyze the mechanism of the attack that explains its success and lack of success for specific speech signals. We can analyze the \textit{saliency} of the input audio to determine the sensitivity of the Whisper's predictions to different parts of the input audio. The frames in the input audio that the transcription is most sensitive to are the parts of the audio that dominate Whisper's decisions. For a model, $\mathcal F(\cdot)$ defined in Equation \ref{eqn:dist}, we can define the $m$-th saliency of the universal adversarial audio segment, $\hat{\tilde{\mathbf x}}$, as the gradient of the $m$-th transcribed token, $\tilde{y}^*_m$,
\begin{equation}
    \tilde s_m = \left|\left|\nabla_{\hat{\tilde{\mathbf x}}}\left[\mathcal F(\hat{\tilde{\mathbf x}}\oplus\mathbf x, \mathbf y^*_{<m};\theta)_{\tilde{y}^*_m}\right ]\right|\right|_2.
\end{equation}
Equivalently we can define the saliency of the natural speech signal, $\mathbf x$ as,
\begin{equation} 
    s_m = \left|\left|\nabla_{\mathbf x}\left[\mathcal F(\hat{\tilde{\mathbf x}}\oplus\mathbf x, \mathbf y^*_{<m};\theta)_{y^*_m}\right ]\right|\right|_2.
\end{equation}
As we are interested primarily in the first generated token, we set $m=1$ in our analysis.
\section{Experiments}


\subsection{Experimental Setup}

\paragraph{Data.} Results are reported across five diverse and popular speech recognition datasets: LibriSpeech (LBS)~\citep{panayotov2015librispeech}, TED-LIUM3 (TED)~\citep{hernandez2018ted}, MGB~\citep{bell2015mgb}; Artie Bias (Artie)~\citep{meyer2020artie} and Fleurs~\citep{fleurs2022arxiv}. Details for each dataset are provided in Section \ref{sec:app-data}. The universal acoustic attack segment is learnt using the development split of the LBS dataset. The attack is then evaluated on the LBS test split and to measure the transferability of the attack it is also evaluated on the other datasets (TED-LIUM3, MGB and Artie Bias). The attack is evaluated for task transferability by also evaluating on speech transcription and speech translation tasks using the Fleurs dataset test splits.
\paragraph{Models.} Experimental results are given for the family of Whisper ASR models \cite{radford2023robust}. Model details and their performance (Word Error Rate) on the datasets have been provided for reference in Appendix \ref{sec:app-models}.
\paragraph{Attack Train Configuration.} The universal acoustic prepend attack segment is trained on the LibriSpeech development split. The attack segment is trained as per Equation \ref{eqn:universal}, where it is prepended to speech samples in the raw audio space. The attack segment length is set to be 0.64 seconds and its maximum amplitude to $\epsilon=0.02$, to satisfy the constraint of Equation \ref{eqn:const}. Further Hyperparameter settings for training the universal acoustic attack segment are given in Appendix \ref{sec:app-config}.


\subsection{Results}

\paragraph{Universal Acoustic Prepend Attack.} The universal prepend attack segment is trained (on the LBS development split) to make the ASR model generate only an $\texttt{<|endoftext|>}$ token, i.e. transcribe nothing. Evaluating on the LBS test-split, Table \ref{tab:direct} gives the percentage of successful attacks, $\varnothing$ and the average sequence length of predicted transcriptions ($\texttt{asl}$) for the different target speech recognition models with the same (per model) trained 0.64-second universal acoustic adversarial segment prepended to every speech sample. A comparison is made to the \textit{no attack} setting, where the speech samples are not modified in any manner. For every target Whisper model, the universal acoustic prepend attack is extremely successful in ensuring the model does not transcribe the speech signals, with the percentage of successful attacks increasing from more than 97\% for the medium models to 99.9\% for the tiny models. Similarly, in all cases the $\texttt{asl}$ is brought to less than 1.0, whereas for the unattacked speech the transcriptions have nearly 18 words on average. We also compare to a random audio segment prepended to the speech samples and we find that this behaves identically to the \textit{no attack} setting, i.e. a random attack cannot `mute' Whisper. Overall, Table \ref{tab:direct} shows that regardless of the model size, a short 0.64-second universal acoustic adversarial audio segment can be prepended (imperceptibly) to almost all speech signals to conceal the contents from Whisper speech recognition models.

\begin{table}[htb!]
    \centering
    \small
    \fontsize{8pt}{8pt}\selectfont
    \begin{tabular}{l|l|cc}
    \toprule
      Model   & Metric & No Attack & Attack \\ \midrule
      
      \multirow{2}{*}{tiny.en} & $\varnothing$ (\%) $\uparrow$ &0.0 & 99.7\\
       & $\texttt{asl}\downarrow$ & 17.9 & 0.06\\ \midrule

      \multirow{2}{*}{tiny} & $\varnothing$ (\%) $\uparrow$ &0.0 & 99.6\\
       & $\texttt{asl}\downarrow$ & 17.9 & 0.04\\ \midrule

      \multirow{2}{*}{base.en} & $\varnothing$ (\%) $\uparrow$ &0.0 & 99.0\\
       & $\texttt{asl}\downarrow$ & 17.8 & 0.20\\ \midrule

      \multirow{2}{*}{base} & $\varnothing$ (\%) $\uparrow$ &0.0 & 99.5\\
       & $\texttt{asl}\downarrow$ & 17.8 & 0.05\\ \midrule

      \multirow{2}{*}{small.en} & $\varnothing$ (\%) $\uparrow$ &0.0 & 98.6\\
       & $\texttt{asl}\downarrow$ & 17.7 & 0.14\\ \midrule

      \multirow{2}{*}{small} & $\varnothing$ (\%) $\uparrow$ &0.0 & 98.7 \\
       & $\texttt{asl}\downarrow$ & 17.3 & 0.15\\ \midrule

      \multirow{2}{*}{medium.en} & $\varnothing$ (\%) $\uparrow$ &0.0 & 99.5\\
       & $\texttt{asl}\downarrow$ & 17.7 & 0.10\\ \midrule

      \multirow{2}{*}{medium} & $\varnothing$ (\%) $\uparrow$ &0.0 & 97.8\\
       & $\texttt{asl}\downarrow$ & 17.8 & 0.56\\ \midrule
    \end{tabular}
    \caption{The percentage of successfully `muted' speech samples, $\varnothing$, where the first generated token is $\texttt{<|endoftext|>}$, and the Average Sequence Length ($\texttt{asl}$) of transcriptions, for the LBS dataset. Results are presented for \textit{no attack}, and for a trained (per model) universal acoustic adversarial attack, where the \textit{same} universal adversarial segment is prepended to each speech sample.}
    \label{tab:direct}
\end{table}

\begin{figure}[htb!]
    \centering
    \includegraphics[width=0.9\linewidth]{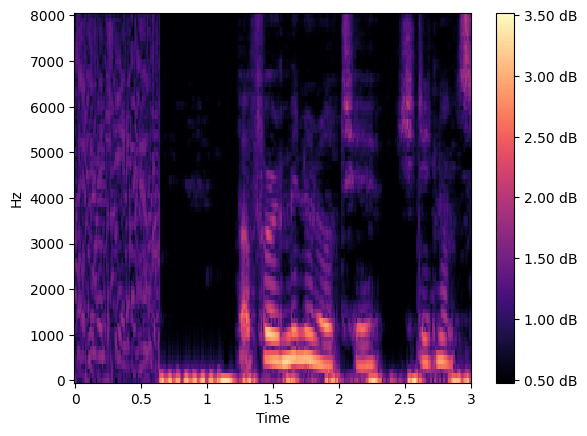}
    \caption{Mel spectrogram of universal acoustic segment (0.64s) prepended to a (truncated) random speech sample from LBS dataset.}
    \label{fig:spect}
\end{figure}

Figure \ref{fig:spect} gives the Mel-spectrogram of a random speech sample from the LBS test set with a 0.64-second universal acoustic adversarial segment prepended to the speech signal (learnt for the medium.en model). This validates that $\epsilon=0.02$ is an appropriate imperceptibility setting as it ensures that the power of the adversarial segment is always less than $\sim 1.50$dB, which is significantly lower than a typical human speech signal in the LBS dataset that can range from 1dB to more than 3.5dB. It is interesting to note that the acoustic adversarial segment covers the full range of frequencies relatively uniformly, which means it is likely to sound similar to static noise to a human listener.

\paragraph{Attack Success Analysis.}
We now investigate the $< 3$\% speech samples for which the universal acoustic attack fails to perfectly mute the Whisper model, i.e., the generated transcription is not of zero-length. Table \ref{tab:asl-failed} gives the average sequence length ($\texttt{asl}$) evaluation of the generated transcripts for the \textit{failed} attack samples (relative to the \textit{successful} samples) for LBS. Interestingly, when there is no adversarial attack, the $\texttt{asl}$ for the failed samples is 2 to 4 times greater than the average $\sim$17 words in the successful samples' transcriptions, suggesting that the universal acoustic attack only struggles to mute the ASR model for longer input speech signals. Further, for these \textit{failed} samples, the attack is still able to reduce the number of generated words significantly (at least two-fold), highlighting that the attack is still effective in muting the ASR model to some extent, although not entirely.

\begin{table}[htb!]
    \centering
    \small
    \fontsize{8pt}{8pt}\selectfont
    \begin{tabular}{l|l|cc}
    \toprule
       Model  & Samples & No Attack & Attack \\ \midrule
       

        \multirow{2}{*}{tiny} & successful & 17.8 & 0.0\\
        & failed & 74.6 & 11.0\\ \midrule






        \multirow{2}{*}{medium} & successful & 17.2 & 0.0\\
        & failed & 43.2 & 25.0\\
        \bottomrule
    \end{tabular}
    \caption{Average Sequence Length ($\texttt{asl}$) of generated transcripts for \textit{successful} attack samples and \textit{failed} attack samples. A successful sample is where the universal acoustic attack causes the Whisper model to generate a zero-length transcription.}
    \label{tab:asl-failed}
\end{table}
A natural follow-up question is then, in what manner does the universal attack shorten the generated transcripts for the \textit{failed} samples, i.e., is it simple truncation or is the model generating other tokens unrelated to the original speech signal. Table \ref{tab:wer} gives the breakdown of the word error rate (WER) contributions from insertions, deletions and substitutions for the \textit{failed} samples, where the word error rate is computed between the predicted \textit{no attack} transcriptions and the predicted \textit{attack} transcriptions. We observe that the attack causes no significant change in the transcriptions other than deletions, demonstrating the attack is behaving as desired in attempting to discourage speech transcription. Overall, this analysis shows that even for the few samples ($< 3$\%) that the universal attack is not able to perfectly mute the ASR model, the attack is still able to significantly reduce the transcription length.

\begin{table}[htb!]
    \centering
    \small
    \fontsize{8pt}{8pt}\selectfont
    \begin{tabular}{l|c|ccc}
    \toprule
       Model  & WER & INS & DEL & SUB \\ \midrule
        tiny & 88.38 & 0.36 & 85.40 & 2.29\\
        medium & 50.76 & 2.70 & 43.75 & 2.94\\
        \bottomrule
    \end{tabular}
    \caption{Word Error Rate (WER) and breakdown (insertions, deletions and substitutions) between the transcript generated with no attack and the transcript generated with the universal attack, for the \textit{failed} attack samples only. A  failed sample is where the universal attack is unable to make Whisper generate a zero-length transcription.}
    \label{tab:wer}
\end{table}

\paragraph{Saliency Analysis.} Section \ref{sec:saliency} describes saliency as a tool to measure the sensitivity of the ASR model to the adversarial and the natural speech segments of the input audio. The average saliencies for the LBS dataset are given in Table \ref{tab:saliency}, with a comparison for the successful attack samples and the failed attack samples. It is clear that a successful attack results in the ASR model being significantly more sensitive to the adversarial segment, and conversely more sensitive to the speech signal when the attack fails. This demonstrates that the universal acoustic attack is operating as intended, as a successful attack encourages the model to attend more to the acoustic realization of the $\texttt{<|endoftext|>}$ special token (the adversarial audio segment).~\footnote{Appendix \ref{sec:app-sal-plots} illustrates the frame-level saliency.} It is also interesting to note that for successful attack samples the saliency is significantly higher, suggesting that success of the adversarial attack is very dependent on the exact learnt universal adversarial segment.

\begin{table}[htb!]
    \centering
    \small
    \fontsize{8pt}{8pt}\selectfont
    \begin{tabular}{l|l|cc}
    \toprule
        Model &  Samples & Adv, $\tilde s$ & Speech, $s$\\ \midrule
        

        \multirow{2}{*}{tiny} & successful  & 835 & 4.80\\
        & failed  & 101 & 192\\ \midrule






        \multirow{2}{*}{medium} & successful  & 3371 & 143\\
        & failed  & 314 & 803\\
        \bottomrule
    \end{tabular}
    \caption{Average saliency for the adversarial segment and speech segment (across LBS dataset) for successful and failed samples. A successful sample is where the universal attack causes Whisper to generate a zero-length transcription.}
    \label{tab:saliency}
\end{table}

\paragraph{Attack Transferability.} The universal attack segment has been trained on a specific domain of data (LBS data) and there is a risk that the attack may not necessarily transfer to different, distributionally shifted speech domains. Therefore, in this section we investigate the impact of transferring the 0.64-second universal acoustic adversarial segment to different unseen (during training of the attack) datasets, representing a diverse range of domain distributional shifts. Table \ref{tab:data-transfer} presents the results. For all models and datasets, the universal acoustic attack is able to continue muting the Whisper models for more than 90\% of samples. Although this is slightly lower than 97\% success rate for the in-domain LBS dataset, 90\% is still a significant success rate, suggesting that the adversarial segment truly represents an acoustic realization of the $\texttt{<|endoftext|>}$ token, which universally prevents the transcription of different speech domains.

\begin{table}[htb!]
    \centering
    \small
    \fontsize{8pt}{8pt}\selectfont
    \begin{tabular}{l|l|c|ccc}
    \toprule
         & Metric & \textit{LBS} & TED & MGB & Artie\\ \midrule\midrule
         
        \multirow{2}{*}{\textit{Ref}} 
        & $\varnothing$ (\%)  & \textit{0.0}& \textit{0.0} &\textit{0.0} & \textit{0.0}\\
        & $\texttt{asl}$ & \textit{17.8} 
        &\textit{24.4} & \textit{8.9} & \textit{8.6}\\\midrule \midrule

        \multirow{2}{*}{tiny.en} 
        & $\varnothing$ (\%)  & 99.7 & 99.9 & 99.9 & 100.0\\ & $\texttt{asl}$ & 0.06 & 0.01 & 0.01 & 0.00\\\midrule
        
        \multirow{2}{*}{tiny} 
        & $\varnothing$ (\%)  & 99.6 & 99.0 & 99.3 & 99.2\\ & $\texttt{asl}$ & 0.04 & 0.56 & 0.10 & 0.03\\\midrule

        \multirow{2}{*}{base.en} 
        & $\varnothing$ (\%) & 99.0 & 98.8 & 99.0 & 99.3\\ & $\texttt{asl}$ & 0.20 & 0.32 & 0.09 & 0.03\\\midrule

        \multirow{2}{*}{base} 
        & $\varnothing$ (\%) & 99.5 & 99.9 & 99.5 & 97.4\\ & $\texttt{asl}$ & 0.05 & 0.01 & 0.09 & 0.17\\\midrule

        \multirow{2}{*}{small.en} 
        & $\varnothing$ (\%) & 98.6 & 93.1 & 98.3 & 92.4\\ & $\texttt{asl}$ & 0.14 & 1.71 & 0.20 & 0.49\\\midrule

        \multirow{2}{*}{small} 
        &$\varnothing$ (\%)  & 98.7 & 99.5 & 93.5 & 97.0\\ & $\texttt{asl}$ & 0.15 & 0.21 & 0.43 & 0.16\\\midrule

        \multirow{2}{*}{medium.en} 
        & $\varnothing$ (\%) & 99.5 & 99.8 & 99.7 & 99.7\\ & $\texttt{asl}$ & 0.10 & 0.01 & 0.01 & 0.03\\\midrule

        \multirow{2}{*}{medium} 
        & $\varnothing$ (\%) & 97.8 & 95.2 & 96.4 & 96.9\\ 
        & $\texttt{asl}$ & 0.56 & 1.05 & 0.29 & 0.24\\
        \bottomrule
    \end{tabular}
    \caption{Attack transferability across datasets: the percentage of successfully `muted' speech samples, $\varnothing$, and the Average Sequence Length ($\texttt{asl}$) of generated transcripts with the universal acoustic attack learnt on LBS and evaluated on other datasets. \textit{Ref} is the average reference transcription length.}
    \label{tab:data-transfer}
\end{table}

Beyond transferability across data distributions, we also investigate how well the universal acoustic adversarial attacks transfer across different speech processing tasks. As the multilingual Whisper models can be instructed to perform transcription or speech translation, we evaluate how effective the adversarial segment (trained on Whisper for transcription) is in muting Whisper when used for speech translation. Table \ref{tab:transfer-task} presents attack results for speech translation from French (fr), German (de), Russian (ru) and Korean (ko) to English, from the Fleurs dataset. Two main trends can be identified. First, the attack transfers extremely well for the smaller Whisper models, with attack success rate greater than 94\%, but for the larger models the success rate can drop to less than even 20\%. Second, it appears that the `further' the source language from English, the lower the success rate, e.g., the attack transfers better for French than Korean in general. 

\begin{table}[htb!]
    \centering
    \small
    \fontsize{8pt}{8pt}\selectfont
    \begin{tabular}{l|l|cccc}
    \toprule
        Model & Metric &fr & de & ru & ko \\ \midrule\midrule
        
        \textit{Ref} & $\varnothing$ (\%) & \textit{0.0}  & \textit{0.0} & \textit{0.0} & \textit{0.0}\\
        & $\texttt{asl}$ & \textit{25.3}  & \textit{21.5} & \textit{19.3}  & \textit{14.7}\\ \midrule\midrule

    tiny & $\varnothing$ (\%) & 99.9 & 94.6 & 96.8 & 94.2\\
    & $\texttt{asl}$ & 0.00 & 0.82 & 0.85 & 1.09\\ \midrule

    base & $\varnothing$ (\%) & 73.1 & 70.0 & 34.1 & 7.9\\
    & $\texttt{asl}$ & 6.42 & 6.20 & 13.05 & 8.03\\ \midrule

    small & $\varnothing$ (\%) & 53.4 & 59.1 & 39.2 & 65.7\\
    & $\texttt{asl}$ & 5.01 & 4.45 & 6.11 & 1.68\\ \midrule

    medium & $\varnothing$ (\%) & 10.5 & 50.7 & 21.7 & 15.5\\
    & $\texttt{asl}$ & 13.04 & 4.44 & 14.46 & 8.18\\ 
    \bottomrule
    \end{tabular}
    \caption{Attack transferability across tasks: the percentage of successfully `muted' speech samples, $\varnothing$, and the Average Sequence Length ($\texttt{asl}$) of generated transcripts with the universal acoustic adversarial attack learnt on LBS for the task of \textit{transcription} and evaluated on the Fleurs dataset for the task of speech \textit{translation} to English. Results are presented for the multi-lingual Whisper models.}
    \label{tab:transfer-task}
\end{table}

Next, we explore transferability of the attack across different Whisper models: this is explored analytically and empirically in Appendix \ref{sec:app-model-transfer}. The key finding is that certain attacks can be trained to transfer across models, but due to fundamental differences in the acoustic representation of the $\texttt{<|endoftext|>}$ token for different models, it is unlikely a muting attack will naively transfer to unseen models.

\begin{figure}
    \centering
    \includegraphics[width=0.8\linewidth]{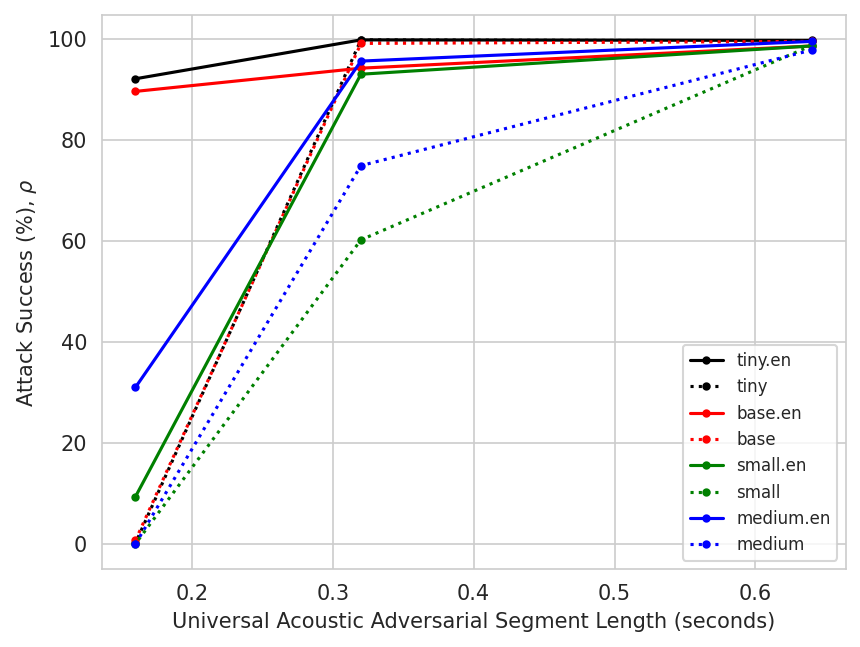}
    \caption{Ablation on the universal acoustic adversarial attack segment length.}
    \label{fig:ablation-T}
\end{figure}

\begin{figure}
    \centering
    \includegraphics[width=0.8\linewidth]{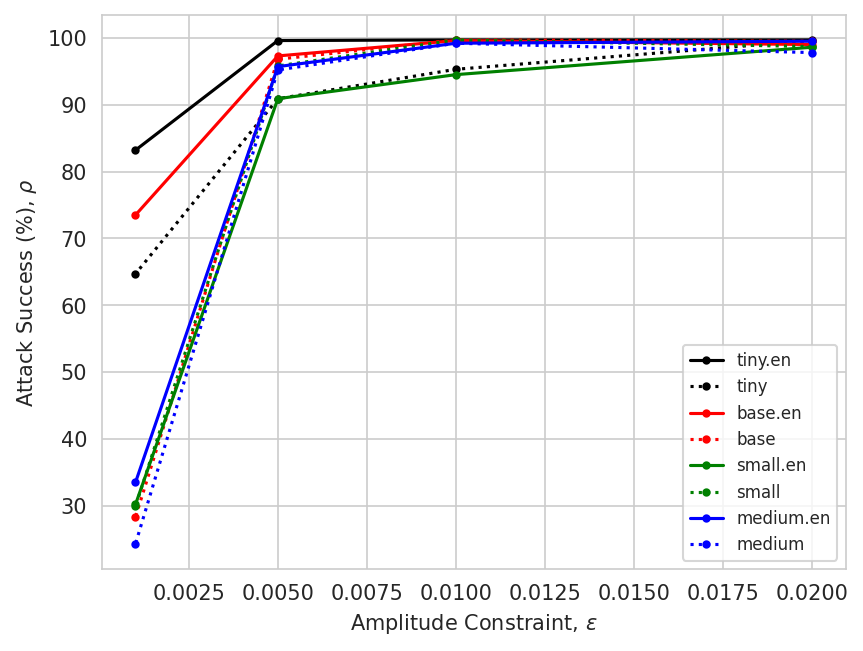}
    \caption{Ablation on the universal acoustic adversarial attack amplitude constraint, $\epsilon$.}
    \label{fig:ablation-e}
\end{figure}

\paragraph{Ablations on Imperceptibility.} 
In this section we explore how much stricter imperceptibility constraints can be made during the training of the universal acoustic attack segments. Figure \ref{fig:ablation-T} shows how the attack success percentage, $\varnothing$ (successfully mute Whisper) changes as the audio segment length is decreased from 0.64-seconds. The larger a model, the greater the decay in attack success. Further, the multi-lingual models tend to have a much greater decay than their English-only counterparts, with the attack success rate reaching near 0\% for every multi-lingual model for a segment of 16-seconds. Figure \ref{fig:ablation-e} equivalently presents the impact of reducing the maximum amplitude, $\epsilon$. A similar trend (although less clear) arises where the larger and the multi-lingual variants of the models have a greater drop in success rate with a smaller $\epsilon$. The relative robustness of the multi-lingual and larger models in extremely constrained attack settings can perhaps be explained simply by the fact these models have been trained on more data and thus it is more difficult to find a universal realization of the $\texttt{<|endoftext|>}$ token.

\section{Conclusion}

This work proposes a highly effective and practical method for `muting' Whisper models, achieving a success rate of over 97\%. A universal 0.64-second adversarial audio segment is trained to represent an acoustic realization of the $\texttt{<|endoftext|>}$ token used by Whisper, such that when this audio segment is prepended to any speech signal, Whisper does not transcribe the speech, i.e., the model is `muted'. Moreover, this universal acoustic adversarial segment transfers across different data distributions and can even transfer to different speech processing tasks. While this result offers a potential for speech privacy protection, it does also reveal the critical security implications of foundation models' susceptibility to adversarial attacks. As speech processing systems continue to develop, addressing these vulnerabilities is an important direction for future research.

\section{Limitations}

We identify the following potential limitations of our work:
\begin{itemize}
    \item The scope of this work covers specifically Transformer-based Automatic Speech Recognition (ASR) systems, such as Whisper. However, due to the recent popularity and performance of Whisper for ASR, this scope is highly relevant for a large number of modern speech processing applications.
    
    \item We demonstrate that the universal adversarial segment can transfer well across different data distributions and even sometimes languages. It would be useful for future work to explore the impact on transferability as specific dimensions of distributional shift are varied in a controlled manner, e.g. amplitude of speech (long-distance vs close-distance audio); level of background noise; or even recording conditions.
    
    \item The universal adversarial attack, although very effective, it is Whisper model specific. This is of course very much expected as each model has a very different audio-space representation. We discuss this in greater detail in Appendix \ref{sec:app-model-transfer}. Although we demonstrate that we can learn a universal attack that is effective for more than one Whisper model (by considering multiple models during training), a defence in the future could be to simply transcribe the text using multiple diverse models. However, we argue that this defence is not only expensive due to linear inference scaling costs, but is extremely uncommon in currently deployed ASR systems - it is more common to use a single ASR system. Hence, if a Whisper model is used for ASR, then an adversary can use the universal acoustic adversarial segment from this work to mute the model.

    \item This work focuses on developing an adversarial attack method to mute the Whisper model. However, we do not explore detection or defence approaches explicitly. This is a research area for future work. However, we also emphasize that it is currently very uncommon in many real-world deployed ASR settings to perform any form of adversarial detection. Therefore, one primary aim of this work is to raise awareness around the vulnerability of Whisper ASR systems to muting universal adversarial attacks. We hope this encourages future research in defence methods where required. Note that our proposed muting adversarial attack method can also be used positively by users to protect the privacy of their audio content.
\end{itemize}

\section{Risks and Ethics}

This work proposes a method to learn a universal acoustic adversarial attack, where a 0.64-second audio segment can be prepended to any speech signal and mute Whisper models. There is the risk that this method could be used by an adversary to conceal the content of speech signals from speech moderation systems. However, we argue the aim of this work is to raise awareness around the vulnerability to such muting adversarial attacks of Whisper ASR models that have been deployed across many speech processing applications. By raising this issue, we hope to encourage the research community to develop methods that improve the robustness and reliability of existing and future ASR systems. Further, the adversarial attack method proposed in this work can also be used constructively by users in speech privacy settings, where it is important to protect the content of audio from malicious actors. On the whole, this research contributes to the rich adversarial attack literature to encourage the further development of safe models.


\bibliography{custom}

\begin{thebibliography}{36}
\expandafter\ifx\csname natexlab\endcsname\relax\def\natexlab#1{#1}\fi

\bibitem[{Abdullah et~al.(2019)Abdullah, Garcia, Peeters, Traynor, Butler, and Wilson}]{abdullah2019practical}
Hadi Abdullah, Washington Garcia, Christian Peeters, Patrick Traynor, Kevin R.~B. Butler, and Joseph Wilson. 2019.
\newblock \href {http://arxiv.org/abs/1904.05734} {Practical hidden voice attacks against speech and speaker recognition systems}.

\bibitem[{Alzantot et~al.(2018)Alzantot, Balaji, and Srivastava}]{DBLP:journals/corr/abs-1801-00554}
Moustafa Alzantot, Bharathan Balaji, and Mani~B. Srivastava. 2018.
\newblock \href {http://arxiv.org/abs/1801.00554} {Did you hear that? adversarial examples against automatic speech recognition}.
\newblock \emph{CoRR}, abs/1801.00554.

\bibitem[{Ardila et~al.(2020)Ardila, Branson, Davis, Kohler, Meyer, Henretty, Morais, Saunders, Tyers, and Weber}]{ardila2020common}
Rosana Ardila, Megan Branson, Kelly Davis, Michael Kohler, Josh Meyer, Michael Henretty, Reuben Morais, Lindsay Saunders, Francis Tyers, and Gregor Weber. 2020.
\newblock {Common Voice: A Massively-Multilingual Speech Corpus}.
\newblock In \emph{Proceedings of the Twelfth Language Resources and Evaluation Conference}, pages 4218--4222.

\bibitem[{Bell et~al.(2015)Bell, Gales, Hain, Kilgour, Lanchantin, Liu, McParland, Renals, Saz, Wester et~al.}]{bell2015mgb}
Peter Bell, Mark~JF Gales, Thomas Hain, Jonathan Kilgour, Pierre Lanchantin, Xunying Liu, Andrew McParland, Steve Renals, Oscar Saz, Mirjam Wester, et~al. 2015.
\newblock {The MGB challenge: Evaluating multi-genre broadcast media recognition}.
\newblock In \emph{2015 IEEE Workshop on Automatic Speech Recognition and Understanding (ASRU)}, pages 687--693. IEEE.

\bibitem[{Carlini et~al.(2016)Carlini, Mishra, Vaidya, Zhang, Sherr, Shields, Wagner, and Zhou}]{197215}
Nicholas Carlini, Pratyush Mishra, Tavish Vaidya, Yuankai Zhang, Micah Sherr, Clay Shields, David Wagner, and Wenchao Zhou. 2016.
\newblock \href {https://www.usenix.org/conference/usenixsecurity16/technical-sessions/presentation/carlini} {Hidden voice commands}.
\newblock In \emph{25th USENIX Security Symposium (USENIX Security 16)}, pages 513--530, Austin, TX. USENIX Association.

\bibitem[{Carlini and Wagner(2018)}]{DBLP:journals/corr/abs-1801-01944}
Nicholas Carlini and David~A. Wagner. 2018.
\newblock \href {http://arxiv.org/abs/1801.01944} {Audio adversarial examples: Targeted attacks on speech-to-text}.
\newblock \emph{CoRR}, abs/1801.01944.

\bibitem[{Chen et~al.(2020)Chen, Yuan, Zhang, Zhao, Zhang, Chen, and Wang}]{247642}
Yuxuan Chen, Xuejing Yuan, Jiangshan Zhang, Yue Zhao, Shengzhi Zhang, Kai Chen, and XiaoFeng Wang. 2020.
\newblock \href {https://www.usenix.org/conference/usenixsecurity20/presentation/chen-yuxuan} {{Devil{\textquoteright}s} whisper: A general approach for physical adversarial attacks against commercial black-box speech recognition devices}.
\newblock In \emph{29th USENIX Security Symposium (USENIX Security 20)}, pages 2667--2684. USENIX Association.

\bibitem[{Cheng et~al.(2024)Cheng, Wu, Hong, Ba, Lin, Lu, and Ren}]{10037237}
P.~Cheng, Y.~Wu, Y.~Hong, Z.~Ba, F.~Lin, L.~Lu, and K.~Ren. 2024.
\newblock \href {https://doi.org/10.1109/TDSC.2023.3242292} {Uniap: Protecting speech privacy with non-targeted universal adversarial perturbations}.
\newblock \emph{IEEE Transactions on Dependable and Secure Computing}, 21(01):31--46.

\bibitem[{Cisse et~al.(2017)Cisse, Adi, Neverova, and Keshet}]{cisse2017houdini}
Moustapha Cisse, Yossi Adi, Natalia Neverova, and Joseph Keshet. 2017.
\newblock \href {http://arxiv.org/abs/1707.05373} {Houdini: Fooling deep structured prediction models}.

\bibitem[{Conneau et~al.(2022)Conneau, Ma, Khanuja, Zhang, Axelrod, Dalmia, Riesa, Rivera, and Bapna}]{fleurs2022arxiv}
Alexis Conneau, Min Ma, Simran Khanuja, Yu~Zhang, Vera Axelrod, Siddharth Dalmia, Jason Riesa, Clara Rivera, and Ankur Bapna. 2022.
\newblock \href {https://arxiv.org/abs/2205.12446} {Fleurs: Few-shot learning evaluation of universal representations of speech}.
\newblock \emph{arXiv preprint arXiv:2205.12446}.

\bibitem[{Das et~al.(2018)Das, Shanbhogue, Chen, Chen, Kounavis, and Chau}]{DBLP:journals/corr/abs-1805-11852}
Nilaksh Das, Madhuri Shanbhogue, Shang{-}Tse Chen, Li~Chen, Michael~E. Kounavis, and Duen~Horng Chau. 2018.
\newblock \href {http://arxiv.org/abs/1805.11852} {{ADAGIO:} interactive experimentation with adversarial attack and defense for audio}.
\newblock \emph{CoRR}, abs/1805.11852.

\bibitem[{Du et~al.(2019)Du, Ji, Li, Gu, Wang, and Beyah}]{du2019sirenattack}
Tianyu Du, Shouling Ji, Jinfeng Li, Qinchen Gu, Ting Wang, and Raheem Beyah. 2019.
\newblock \href {http://arxiv.org/abs/1901.07846} {Sirenattack: Generating adversarial audio for end-to-end acoustic systems}.

\bibitem[{Fan et~al.(2020)Fan, Li, Jiang, Xu, and Lu}]{9348184}
Wenshu Fan, Hongwei Li, Wenbo Jiang, Guowen Xu, and Rongxing Lu. 2020.
\newblock \href {https://doi.org/10.1109/GLOBECOM42002.2020.9348184} {A practical black-box attack against autonomous speech recognition model}.
\newblock In \emph{GLOBECOM 2020 - 2020 IEEE Global Communications Conference}, pages 1--6.

\bibitem[{Gong and Poellabauer(2017)}]{DBLP:journals/corr/abs-1711-03280}
Yuan Gong and Christian Poellabauer. 2017.
\newblock \href {http://arxiv.org/abs/1711.03280} {Crafting adversarial examples for speech paralinguistics applications}.
\newblock \emph{CoRR}, abs/1711.03280.

\bibitem[{Hernandez et~al.(2018)Hernandez, Nguyen, Ghannay, Tomashenko, and Esteve}]{hernandez2018ted}
Fran{\c{c}}ois Hernandez, Vincent Nguyen, Sahar Ghannay, Natalia Tomashenko, and Yannick Esteve. 2018.
\newblock {TED-LIUM 3: Twice as much data and corpus repartition for experiments on speaker adaptation}.
\newblock In \emph{Speech and Computer: 20th International Conference, SPECOM 2018, Leipzig, Germany, September 18--22, 2018, Proceedings 20}, pages 198--208. Springer.

\bibitem[{Khare et~al.(2019)Khare, Aralikatte, and Mani}]{khare2019adversarial}
Shreya Khare, Rahul Aralikatte, and Senthil Mani. 2019.
\newblock \href {http://arxiv.org/abs/1811.01312} {Adversarial black-box attacks on automatic speech recognition systems using multi-objective evolutionary optimization}.

\bibitem[{Li et~al.(2020)Li, Wu, Liu, Chen, and Yuan}]{10.1145-3372297.3423348}
Zhuohang Li, Yi~Wu, Jian Liu, Yingying Chen, and Bo~Yuan. 2020.
\newblock \href {https://doi.org/10.1145/3372297.3423348} {Advpulse: Universal, synchronization-free, and targeted audio adversarial attacks via subsecond perturbations}.
\newblock In \emph{Proceedings of the 2020 ACM SIGSAC Conference on Computer and Communications Security}, CCS '20, page 1121–1134, New York, NY, USA. Association for Computing Machinery.

\bibitem[{Lu et~al.(2021)Lu, Han, Zhang, and Cao}]{lu2021exploring}
Zhiyun Lu, Wei Han, Yu~Zhang, and Liangliang Cao. 2021.
\newblock \href {http://arxiv.org/abs/2104.02757} {Exploring targeted universal adversarial perturbations to end-to-end asr models}.

\bibitem[{Ma et~al.(2021)Ma, Chen, and Yong}]{ma2021simulating}
Chen Ma, Li~Chen, and Jun-Hai Yong. 2021.
\newblock \href {http://arxiv.org/abs/2009.00960} {Simulating unknown target models for query-efficient black-box attacks}.

\bibitem[{MacAvaney et~al.(2019)MacAvaney, Yao, Yang, Russell, Goharian, and Frieder}]{macavaney2019hate}
Sean MacAvaney, Hao-Ren Yao, Eugene Yang, Katina Russell, Nazli Goharian, and Ophir Frieder. 2019.
\newblock Hate speech detection: Challenges and solutions.
\newblock \emph{PloS one}, 14(8):e0221152.

\bibitem[{Madry et~al.(2019)Madry, Makelov, Schmidt, Tsipras, and Vladu}]{madry2019deep}
Aleksander Madry, Aleksandar Makelov, Ludwig Schmidt, Dimitris Tsipras, and Adrian Vladu. 2019.
\newblock \href {http://arxiv.org/abs/1706.06083} {Towards deep learning models resistant to adversarial attacks}.

\bibitem[{Meyer et~al.(2020)Meyer, Rauchenstein, Eisenberg, and Howell}]{meyer2020artie}
Josh Meyer, Lindy Rauchenstein, Joshua~D Eisenberg, and Nicholas Howell. 2020.
\newblock {Artie bias corpus: An open dataset for detecting demographic bias in speech applications}.
\newblock In \emph{Proceedings of the twelfth language resources and evaluation conference}, pages 6462--6468.

\bibitem[{Neekhara et~al.(2019)Neekhara, Hussain, Pandey, Dubnov, McAuley, and Koushanfar}]{DBLP:journals/corr/abs-1905-03828}
Paarth Neekhara, Shehzeen Hussain, Prakhar Pandey, Shlomo Dubnov, Julian~J. McAuley, and Farinaz Koushanfar. 2019.
\newblock \href {http://arxiv.org/abs/1905.03828} {Universal adversarial perturbations for speech recognition systems}.
\newblock \emph{CoRR}, abs/1905.03828.

\bibitem[{Olivier and Raj(2023)}]{olivier2023kind}
Raphael Olivier and Bhiksha Raj. 2023.
\newblock \href {http://arxiv.org/abs/2210.17316} {There is more than one kind of robustness: Fooling whisper with adversarial examples}.

\bibitem[{Panayotov et~al.(2015)Panayotov, Chen, Povey, and Khudanpur}]{panayotov2015librispeech}
Vassil Panayotov, Guoguo Chen, Daniel Povey, and Sanjeev Khudanpur. 2015.
\newblock {Librispeech: an ASR corpus based on public domain audio books}.
\newblock In \emph{2015 IEEE international conference on acoustics, speech and signal processing (ICASSP)}, pages 5206--5210. IEEE.

\bibitem[{Qin et~al.(2019)Qin, Carlini, Goodfellow, Cottrell, and Raffel}]{qin2019imperceptible}
Yao Qin, Nicholas Carlini, Ian Goodfellow, Garrison Cottrell, and Colin Raffel. 2019.
\newblock \href {http://arxiv.org/abs/1903.10346} {Imperceptible, robust, and targeted adversarial examples for automatic speech recognition}.

\bibitem[{Radford et~al.(2022)Radford, Kim, Xu, Brockman, McLeavey, and Sutskever}]{radford2022robust}
Alec Radford, Jong~Wook Kim, Tao Xu, Greg Brockman, Christine McLeavey, and Ilya Sutskever. 2022.
\newblock \href {http://arxiv.org/abs/2212.04356} {Robust speech recognition via large-scale weak supervision}.

\bibitem[{Radford et~al.(2023)Radford, Kim, Xu, Brockman, McLeavey, and Sutskever}]{radford2023robust}
Alec Radford, Jong~Wook Kim, Tao Xu, Greg Brockman, Christine McLeavey, and Ilya Sutskever. 2023.
\newblock {Robust speech recognition via large-scale weak supervision}.
\newblock In \emph{International Conference on Machine Learning}, pages 28492--28518. PMLR.

\bibitem[{Raina et~al.(2020)Raina, Gales, and Knill}]{raina_gales_knill_2020}
V~Raina, MJF Gales, and K~Knill. 2020.
\newblock \href {https://doi.org/10.17863/CAM.63488} {Universal adversarial attacks on spoken language assessment systems}.
\newblock \emph{Interspeech}.

\bibitem[{Sch{\"o}nherr et~al.(2018)Sch{\"o}nherr, Kohls, Zeiler, Holz, and Kolossa}]{Schnherr2018AdversarialAA}
Lea Sch{\"o}nherr, Katharina~Siobhan Kohls, Steffen Zeiler, Thorsten Holz, and Dorothea Kolossa. 2018.
\newblock \href {https://api.semanticscholar.org/CorpusID:52040758} {Adversarial attacks against automatic speech recognition systems via psychoacoustic hiding}.
\newblock \emph{ArXiv}, abs/1808.05665.

\bibitem[{Schönherr et~al.(2018)Schönherr, Kohls, Zeiler, Holz, and Kolossa}]{schönherr2018adversarial}
Lea Schönherr, Katharina Kohls, Steffen Zeiler, Thorsten Holz, and Dorothea Kolossa. 2018.
\newblock \href {http://arxiv.org/abs/1808.05665} {Adversarial attacks against automatic speech recognition systems via psychoacoustic hiding}.

\bibitem[{Taori et~al.(2019)Taori, Kamsetty, Chu, and Vemuri}]{taori2019targeted}
Rohan Taori, Amog Kamsetty, Brenton Chu, and Nikita Vemuri. 2019.
\newblock \href {http://arxiv.org/abs/1805.07820} {Targeted adversarial examples for black box audio systems}.

\bibitem[{Wu and Bhandary(2020)}]{wu2020detection}
Ching~Seh Wu and Unnathi Bhandary. 2020.
\newblock Detection of hate speech in videos using machine learning.
\newblock In \emph{2020 International Conference on Computational Science and Computational Intelligence (CSCI)}, pages 585--590. IEEE.

\bibitem[{Yuan et~al.(2018)Yuan, Chen, Zhao, Long, Liu, Chen, Zhang, Huang, Wang, and Gunter}]{yuan2018commandersong}
Xuejing Yuan, Yuxuan Chen, Yue Zhao, Yunhui Long, Xiaokang Liu, Kai Chen, Shengzhi Zhang, Heqing Huang, Xiaofeng Wang, and Carl~A. Gunter. 2018.
\newblock \href {http://arxiv.org/abs/1801.08535} {Commandersong: A systematic approach for practical adversarial voice recognition}.

\bibitem[{Zhang et~al.(2017)Zhang, Yan, Ji, Zhang, Zhang, and Xu}]{DBLP:journals/corr/abs-1708-09537}
Guoming Zhang, Chen Yan, Xiaoyu Ji, Taimin Zhang, Tianchen Zhang, and Wenyuan Xu. 2017.
\newblock \href {http://arxiv.org/abs/1708.09537} {Dolphinatack: Inaudible voice commands}.
\newblock \emph{CoRR}, abs/1708.09537.

\bibitem[{Zheng et~al.(2021)Zheng, Jiang, Wang, Li, Shen, Wang, Ge, Teng, and Zhang}]{Zheng_2021}
Baolin Zheng, Peipei Jiang, Qian Wang, Qi~Li, Chao Shen, Cong Wang, Yunjie Ge, Qingyang Teng, and Shenyi Zhang. 2021.
\newblock \href {https://doi.org/10.1145/3460120.3485383} {Black-box adversarial attacks on commercial speech platforms with minimal information}.
\newblock In \emph{Proceedings of the 2021 ACM SIGSAC Conference on Computer and Communications Security}, CCS ’21. ACM.

\end{thebibliography}

\appendix


\section{Experimental Details}

This section provides greater detail for the experiments in the main paper.

\subsection{Data} \label{sec:app-data}

 The LibriSpeech dataset~\citep{panayotov2015librispeech} is derived from English audio-books and consists of a total of nearly 1000 hours of audio (and transcriptions). In this work, we use specifically the \textit{dev-other} split (2864 utterances forming 5.3 hours of audio) and the \textit{test-other} split (2939 utterances forming 5.1 hours of audio). The TED-LIUM3 dataset~\citep{hernandez2018ted} is formed from English-language TED talks, where the test split consists of 1155 utterances and 2.6 hours of audio. The Multi-Genre Broadcast (MGB) Challenge~\citep{bell2015mgb}, an evaluation focused on speech recognition, speaker diarization, and `lightly supervised' alignment of BBC TV recordings. The challenge training data covered the whole range of seven weeks BBC TV output across four channels, resulting in about 1,600 hours of broadcast audio. In addition several hundred million words of BBC subtitle text was provided for language modelling. The Artie Bias dataset~\citep{meyer2020artie} is a subset of the Mozilla Common Voice \cite{ardila2020common} corpus, where it was designed to detect demographic bias in speech applications. The test-split used in this work consists of 1712 utterances forming 2.4 hours of audio. The Few-shot Learning Evaluation of Universal Representation of Speech (Fleurs)~\citep{fleurs2022arxiv} is a n-way parallel speech dataset in 102 languages, with 12 hours of speech per language. For this work we evaluate on the test splits of specifically French (fr), German (de), Russian (ru) and Korean (ko).

\subsection{Models} \label{sec:app-models}
Whisper model checkpoints are available in a range of sizes: Whisper tiny (39M parameters); Whisper base (74M); Whisper small (244M); Whisper medium (769M); and Whisper large (1.55B parameters). The Whisper models are available as English-only (en) or multilingual models. Whisper large is only available as a multilingual model. The Whisper models can be prompted to do speech recognition, voice activity detection, as well as speech translation and language identification for the multi-lingual model variants. This work considers a range of sizes of Whisper models for speech recognition and the multilingual versions are also evaluated for speech translation: tiny(.en), base(.en), small(.en) and medium(.en). The performance of each model, measured by the Word Error Rate (WER), for each dataset is given in Table \ref{tab:perf}. Further, in all experiments we use Whisper's default decoding strategy with a beam size of 5.

\begin{table}[htb!]
    \centering
    \small
    \fontsize{8pt}{8pt}\selectfont
    \begin{tabular}{lcccc}
    \toprule
      Model   & LBS & TED & MGB & Artie \\ \midrule
        tiny.en & 12.8 & 5.4 & 24.5 &18.4\\
        tiny & 15.0 & 6.3 & 29.5 &20.8\\
        base.en & 9.6 &4.6 &19.7& 13.2\\
        base & 11.0 & 5.0 &22.0& 15.3\\
        small.en & 6.7 &4.3 &14.1& 9.2\\
        small & 7.2 & 4.3 &15.0 & 9.3\\
        medium.en & 5.7 & 4.3 & 12.4 &7.4\\
        medium & 5.6 & 4.0 & 12.3 & 6.7\\
        \bottomrule
    \end{tabular}
    \caption{Whisper Model Performance - Word Error Rate (WER), \%.}
    \label{tab:perf}
\end{table}

\subsection{Attack Train Configuration} \label{sec:app-config}

Gradient descent based training is used to learn the acoustic adversarial segment to minimize the loss, which is defined as the negative of the log-likelihood of the probability defined in Equation \ref{eqn:universal}. Note that the Whisper model weights are frozen. The training hyperparamaters for learning the adversarial attack segment are: the use of an AdamW optimizer; a learning rate of 1e-3; a batch size of 16 (apart from medium(.en), where a batch size of 4 was used); and parameter clipping in each gradient step, to clamp the learnt attack segment values of each frame to a maximum absolute value of $\epsilon=0.02$ to satisfy the imperceptibility constraint, as given in Equation \ref{eqn:const}. The larger the target Whisper model, the greater the number of training epochs are required to guarantee a successful universal attack segment. The following number of training epochs are used for each Whisper model: tiny(.en) (40 epochs); base(.en) (40 epochs); small(.en) (120 epochs); and medium(.en) (160 epochs). Note that for the base and base.en models, runs over 2 seeds and 3 seeds respectively were required to find a universal adversarial audio segment that was sufficiently powerful (the seed controls the initialization of the adversarial audio segment during its training). Further note that it is empirically observed that increasing the number of training epochs only increases the strength of the universal attack - there is no risk of overfitting, which is perhaps expected as there are so few values being learnt for the universal attack segments. 

In typical training setups, there is a risk that excessive training steps can lead to overfitting, compromising test-time evaluation. However, when learning the universal prepend attack in this work, this risk does not exist, as the total number of parameters being learnt are only 10,240 parameters for 0.64-second of audio sampled at 16kHz. This is far smaller than the 100s of millions of parameters typically being trained in the Whisper speech recognition models. As a result, we find that the universal prepend attacks learnt in this work transfer perfectly from the development split of the LBS data on which they are trained, to the test split on which they are evaluated, as per the metrics $\varnothing$ and $\texttt{asl}$, used in this paper. 

In the main paper we evaluate the Whisper models in their default setting, where there is no use of the $\texttt{<notimestamps>}$ special token, such that the first generated token by the model is always $\texttt{<|0.0|>}$, and only then the text tokens follow. However, during training/learning of the universal attack, we initialized $\mathbf{y}_0^*$ as $\texttt{<startoftranscript>}$ $\texttt{<language>}$ $\texttt{<task>}$ $\texttt{<notimestamps>}$ and train to predict $y_1=\texttt{<|endoftext|>}$. We find that training the attack with this $\mathbf{y}_0^*$ yields more effective attacks for the multilingual Whisper models. The fact that the attack transfers so well from training time to test time (despite the mismatch in decoder input initialization), suggests that we have learnt a genuine acoustic realization of the $\texttt{<|endoftext|>}$ special token.

A further point to note is that we conducted separate experiments to confirm that when evaluating the adversarial attack, for no sample is the voice activity detector (used as part of Whisper's transcription framework) returning `no speech', i.e., the universal acoustic adversarial segment is a genuine realization of the $\texttt{<|endoftext|>}$ special token. It is unlikely the voice activity detector would ever be activated at evaluation time as during the training of the universal attack segment the internal voice activity detector is not present.

\subsection{Computational Requirements}

Experiments were run on the A100 Nvidia GPU hardware. To learn the 0.64-second universal acoustic adversarial attack using the development split of the LBS dataset, the number of GPU hours vary with the target model size and the number of training epochs used per model. Table \ref{tab:app-compute} summarizes the training epochs (for a successful attack) and the number of subsequent required GPU hours for each model size. Further note that the \textit{medium} models required a maximum batch size of 4 to fit in the GPU RAM, whilst the other models could afford a batch size of 16.

\begin{table}[htb!]
    \centering
    \small
    \fontsize{8pt}{8pt}\selectfont
    \begin{tabular}{l|cc}
    \toprule
       Model  & Epochs & \# GPU hours \\ \midrule
        tiny & 40 & 0.45\\
        base & 40 & 0.92\\
        small & 120 & 2.6\\
        medium & 160 & 8.4\\
        \bottomrule
    \end{tabular}
    \caption{A100 GPU hours to learn a universal acoustic adversarial attack per target model using the development split of the LBS dataset.}
    \label{tab:app-compute}
\end{table}

\subsection{Licensing}
All datasets used are publicly available or specifically approved for experiments in this work (MGB3). Our implementation utilizes the PyTorch 1.12 framework, an open-source library. We observe the MIT license under which the Whisper's code and model weights are released.

\section{Complete Experimental Analysis Results} \label{sec:app-complete}

Experimental results in the main paper are presented for eight Whisper models. However, the results for the \textit{attack success analysis} (Table \ref{tab:asl-failed} and Table \ref{tab:wer}) and the \textit{saliency analysis} (Table \ref{tab:saliency}) are given for only the tiny and medium model. Here we present the full results on all eight different models for completeness. The results maintain the same trends as stated in the analysis in the main paper. The complete \textit{attack success analysis} results are given in Table \ref{tab:app-asl-failed} and Table \ref{tab:app-wer}, whereas the the complete \textit{saliency analysis} results are given in Table \ref{tab:app-saliency}.

\begin{table}[htb!]
    \centering
    \small
    \fontsize{8pt}{8pt}\selectfont
    \begin{tabular}{l|l|cc}
    \toprule
       Model  & Samples & No Attack & Attack \\ \midrule
       
        \multirow{2}{*}{tiny.en} & successful & 17.8 & 0.0\\
        & failed &78.5 & 16.1 \\ \midrule

        \multirow{2}{*}{tiny} & successful & 17.8 & 0.0\\
        & failed & 74.6 & 11.0\\ \midrule

        \multirow{2}{*}{base.en} & successful & 17.5 & 0.0\\
        & failed & 50.4 & 19.4\\ \midrule

        \multirow{2}{*}{base} & successful & 17.6 & 0.0\\
        & failed & 60.2 & 11.4\\ \midrule

        \multirow{2}{*}{small.en} & successful & 17.5 & 0.0\\
        & failed & 31.4 & 10.5\\ \midrule

        \multirow{2}{*}{small} &  successful & 17.1 & 0.0\\
        & failed & 38.7 & 11.7\\ \midrule

        \multirow{2}{*}{medium.en} & successful & 17.4 & 0.0\\
        & failed & 64.8 & 18.9\\ \midrule

        \multirow{2}{*}{medium} & successful & 17.2 & 0.0\\
        & failed & 43.2 & 25.0\\
        \bottomrule
    \end{tabular}
    \caption{Average Sequence Length ($\texttt{asl}$) of generated transcripts for \textit{successful} attack samples and \textit{failed} attack samples. A successful sample is where the universal acoustic attack causes the Whisper model to generate a zero-length transcription (perfectly muted).}
    \label{tab:app-asl-failed}
\end{table}

\begin{table}[htb!]
    \centering
    \small
    \fontsize{8pt}{8pt}\selectfont
    \begin{tabular}{l|c|ccc}
    \toprule
       Model  & WER & INS & DEL & SUB \\ \midrule
        tiny.en & 80.02 & 0.00 & 79.52 & 0.51\\
        tiny & 88.38 & 0.36 & 85.40 & 2.29\\
        base.en & 64.46 & 0.38 & 61.30 & 2.53\\
        base & 89.57 & 1.97 & 81.30 & 4.53\\
        small.en & 75.50 & 0.24 & 66.46 & 8.62\\
        small & 72.95 & 0.40 & 69.02 & 3.23\\
        medium.en & 72.88 & 0.38 & 70.79 & 1.44 \\
        medium & 50.76 & 2.70 & 43.75 & 2.94\\
        \bottomrule
    \end{tabular}
    \caption{Word Error Rate (WER) and breakdown (insertions, deletions and substitutions) between the transcript generated with no attack and the transcript generated with the universal acoustic attack, for the \textit{failed} attack samples only. A  failed sample is where the universal acoustic attack is unable to make Whisper generate a zero-length transcription.}
    \label{tab:app-wer}
\end{table}

\begin{table}[htb!]
    \centering
    \small
    \fontsize{8pt}{8pt}\selectfont
    \begin{tabular}{l|l|cc}
    \toprule
        Model &  Samples & Adv, $\tilde s$ & Speech, $s$\\ \midrule
        
        \multirow{2}{*}{tiny.en} & successful  & $\underset{\pm264}{617}$ & $\underset{\pm15.6}{1.12}$\\
        & failed  & $\underset{\pm97.1}{61.0}$ & $\underset{\pm107}{65.8}$\\ \midrule

        \multirow{2}{*}{tiny} & successful  & $\underset{\pm332}{835}$ & $\underset{\pm49.0}{4.80}$\\
        & failed  & $\underset{\pm33.1}{101}$ & $\underset{\pm517}{192}$\\ \midrule

        \multirow{2}{*}{base.en} & successful  & $\underset{\pm1325}{3527}$ & $\underset{\pm46.8}{6.05}$\\
        & failed  & $\underset{\pm198}{343}$ & $\underset{\pm246}{91.8}$\\ \midrule

        \multirow{2}{*}{base} & successful  & $\underset{\pm1480}{4946}$ & $\underset{\pm140}{13.9}$\\
        & failed & $\underset{\pm183}{483}$ & $\underset{\pm683}{509}$\\ \midrule

        \multirow{2}{*}{small.en} & successful  & $\underset{\pm1263}{4339}$ & $\underset{\pm309}{26.6}$\\
        & failed &  $\underset{\pm308}{727}$ & $\underset{\pm619}{375}$\\ \midrule

        \multirow{2}{*}{small} & successful  & $\underset{\pm1082}{3502}$ & $\underset{\pm102}{23.1}$\\
        & failed  & $\underset{\pm254}{447}$ & $\underset{\pm395}{356}$\\ \midrule

        \multirow{2}{*}{medium.en} & successful  & $\underset{\pm1099}{3205}$ & $\underset{\pm1185}{123}$\\
        & failed &  $\underset{\pm33.4}{114}$ & $\underset{\pm1950}{812}$\\ \midrule

        \multirow{2}{*}{medium} & successful  & $\underset{\pm1254}{3371}$ & $\underset{\pm548}{143}$\\
        & failed  & $\underset{\pm170}{314}$ & $\underset{\pm950}{803}$\\
        \bottomrule
    \end{tabular}
    \caption{Average saliency for the adversarial segment and speech segment (across LBS dataset) for successful and failed samples. A successful sample is where the universal acoustic attack causes the Whisper model to generate a zero-length transcription (perfectly muted).}
    \label{tab:app-saliency}
\end{table}

\newpage
~\newpage
~\newpage

\section{Transferability Across Models} \label{sec:app-model-transfer}

In this section we explore the transferability of the learnt universal acoustic adversarial attack segments across different Whisper models. Table \ref{tab:app-transfer} shows that there is no naive transferability of the adversarial audio segments across models. We next explain this result analytically. Based on the analysis, we further explore empirical methods to try and find adversarial audio segments that could transfer between models.


\begin{table}[htb!]
    \centering
    \small
    \fontsize{8pt}{8pt}\selectfont
    \begin{tabular}{l|l|cc}
    \toprule
        src & tgt & $\varnothing$ (\%) & $\texttt{asl}$\\ \midrule
        tiny & base & 0.0 & 17.8\\
        tiny & small & 0.0 & 17.3\\
        tiny & medium & 0.0 & 17.8\\ \midrule
        medium & small & 0.0 & 17.3\\
        medium & base & 0.0 & 17.8\\
        medium & tiny & 0.0 & 17.9\\
        \bottomrule
    \end{tabular}
    \caption{Transferability of universal acoustic adversarial attack learnt on the source (\textit{src}) model and evaluated on the target (\textit{tgt}) model.}
    \label{tab:app-transfer}
\end{table}

\subsection{Analytically understanding the transferability across models}

Let $\mathbf q^{[k]}$ be the embedding generated by the final layer of the Transformer decoder, to be used to predict the next token (in the case of a muting whisper attack, the first token). For a vocabulary $\mathcal V$, we obtain the logits predicted by the model, $\mathbf y^{[|\mathcal V|]}$ via a projection matrix, $\mathbf W^{[|\mathcal V|\times k]}$~\footnote{The projection matrix $\mathbf W$ is the same as the embedding matrix used at the input to the decoder.},
\begin{equation}
    \mathbf y = \mathbf W\mathbf q,
\end{equation}
where a greedy decoder selects the token, $\hat j$ with the largest logit value,
\begin{equation}
    \hat j = \argmax_{j}\{y_j\}.
\end{equation}
If we define the projection matrix using row vectors,
\begin{equation}
\mathbf{W} =
\begin{bmatrix}
\text{-----}\mathbf{w}_1\text{-----} \\
\text{-----}\mathbf{w}_2\text{-----} \\
\vdots \\
\text{-----}\mathbf{w}_{|\mathcal{V}|}\text{-----}
\end{bmatrix},
\end{equation}
then the greedily selected token can be equivalently selected as,
\begin{equation}
    \hat j = \argmax_j\{\mathbf w_j^T\mathbf q\}
\end{equation}
If the first generated token is $\hat j=r$, then you would expect in a perfect system that the acoustic realization (audio segment), $\mathbf x_r$ of token $r$, when input to the encoder, to give,
\begin{equation}
    \mathbf q\approx \mathbf{w}_{r}
\end{equation}
to maximize its selection for generation. Note that you would expect that if row vectors $\mathbf w_r$ and $\mathbf w_j$ are geometrically close (cosine distance) (i.e. the predicted logit values $y_r$ and $y_j$ are positively correlated), the acoustic realizations $\mathbf x_r$ and $\mathbf x_j$ are similar too, i.e. token $r$ and token $j$ have a similar acoustic sound. We know that certain tokens have \textit{real} acoustic sounds (that are model invariant), e.g., normal words like \textit{zoo}, \textit{boy} and \textit{hi} have real acoustic realizations ($\mathbf x$) that are independent of specific models. Let $\mathcal D$ represent the set of tokens that have a real acoustic sound. Then for a model $\theta$, we can define the relative acoustic position of any token $r$ by considering its similarity to each of these real acoustic tokens.

\begin{equation}
\bm{\rho}(r;\theta) = \left[ \rho_1(r;\theta), \rho_2(r;\theta), \ldots, \rho_{|\mathcal{V}|}(r;\theta) \right]
\end{equation}

\begin{equation}
    \rho_i(r;\theta) = \begin{cases}
\tilde{\mathbf{w}}_i^T \tilde{\mathbf{w}}_r & \text{if } i \in \mathcal{D} \\
\text{null} & \text{if } i \notin \mathcal{D}
\end{cases}
\end{equation}

For a token $d\in\mathcal D$, where $\mathcal D$ represents all those tokens that have real sounds (e.g. normal words like \textit{hello}, \textit{zoo}, etc.), we would expect their relative positions (to other real sounds) to be very consistent across different models (This has been demonstrated in Table \ref{tab:similar}). If we define the difference in acoustic representation for any token $r$ as,
\begin{equation}
    s(r; \theta_m, \theta_n) = \|\bm{\rho}(r; \theta_m) - \bm{\rho}(r; \theta_n)\|_2,
\end{equation}
then for a token $d\in\mathcal D$,
\begin{equation}
\forall d \in \mathcal{D}, \quad s(d; \theta_m, \theta_n) \leq \epsilon,
\end{equation}
where $\epsilon$ is an arbitrarily small value.
\begin{table}[htb!]
    \centering
    \small
    \fontsize{7pt}{7pt}\selectfont
    \begin{tabular}{l|c|l}
    \toprule
        Token & Model & Top 5 closest tokens in $\mathcal D$ as per $\mathbf W$ \\ \midrule
        
        zoo & tiny & Z, j, k, ch, iz\\
        & base & Z, j, iz, ch, k\\
        & small & Z, ch, iz, j, zh\\
        & medium & Z, j, ch, rr, ez\\ \midrule

        boy & tiny & boys, girl, Boy, Bry, NOUN\\
        & base & boys, girl, Boy, Missy, Cameraman\\
        & small & boys, girl, Bry, Justin, Boy\\
        & medium & boys, Boy, moil, ontec\\ \midrule

        hi & tiny & Hi, him, HI, iiii, high\\
        & base & HI, Hi, iiii, Cameraman, Katie\\
        & small & HI, Hi, pleasant, Julia, Hola\\
        & medium & Hi, HI, FFFF, Adam, scream\\ \midrule

        $\texttt{eot}$ & tiny & Male, Pro, Sa, Vict, Cho\\
        & base & Arin, JIN, ELLE, ARRATOR, Jared\\
        & small & WW, pleasant, Gra, Hyun, Missy\\
        & medium & Everyone, sound, Something, Come, Aw\\
        
         \bottomrule
    \end{tabular}
    \caption{Exploring the geometric relationship between embedding matrix tokens in $\mathbf W$. As expected, generally similar sounding words are close together. In some cases, similar domain/meaning words are also close. Note that there can be slight differences to the previous table as some tokens in the other table had a space before them.}
    \label{tab:similar}
\end{table}

The acoustic realization (sound) of the $\texttt{eot}$ token is not known, such that $\texttt{eot}\notin\mathcal D$, as it's not a real acoustic sound. However, we can predict which tokens the acoustic realization $\mathbf x_{\texttt{eot}}$ should be similar to, by considering the geometric position of $\mathbf w_{\texttt{eot}}$ relative to other tokens with a real sound, belonging to $\mathcal D$ - we can compute $\bm{\rho}(\texttt{eot};\theta)$. For there to exist a muting adversarial attack audio segment, $\mathbf x_{\texttt{eot}}(\theta)$ that is transferable across different models, $\theta$, the acoustic realization of $\texttt{eot}$ has to be the same/similar for the different models (the way the acoustic realization of any other real token in $\mathcal D$ is the same for all models). We can thus determine if there exists this true, universal acoustic realization of $\texttt{eot}$ that is the same for all models by observing how consistent its relative position is to tokens in $\mathcal D$, as per $\bm{\rho}$. For a pair of models, $\theta_m$ and $\theta_n$, we expect there to exist a transferable muting adversarial attack, $\mathbf x_{\texttt{eot}}$ if,
\begin{equation}
    s(\texttt{eot}; \theta_m,\theta_n) \leq \tau (\theta_m, \theta_n),
\end{equation} 
where we can define the threshold $\tau$ by considering the typical changes in similarity for other tokens with a real sound (belong to $\mathcal D$) that should have a consistent acoustic sound. We give error for variation by setting the threshold to be two standard deviations above the average change in similarity across models,
\begin{align}
    \tau(\theta_m,\theta_n) =& \mathbb{E}_{r \in \mathcal{D}}[s(r; \theta_m,\theta_n)]\nonumber\\ &+ 2\cdot \sigma_{r \in \mathcal{D}}(s(r; \theta_m,\theta_n))
\end{align}

\subsection{Empirical Evaluation of Model Transferability}
We define the set of real acoustic sounds, $\mathcal D$ as the tokens which begin with any English letter (in roman alphabet) or English numeral (0-9). Table \ref{tab:theoretical-transfer} reports uses the projection matrix, $\mathbf W$ of each Whisper model to determine the potential of the attack transferability. It is interesting to note that there is generally a low chance of model transferability, as the expected acoustic representation of the $\texttt{eot}$ token is far less consistent than that of tokens with a real acoustic sound. These results demonstrate that there is no real audio representation for the $\texttt{<|endoftext|>}$ token, and as a result the attack is unable to find a genuine acoustic realization. Hence, the acoustic realization being learnt is a specific realization of the $\texttt{<|endoftext|>}$ token of the target model.

\begin{table}[htb!]
    \centering
    \small
    \fontsize{8pt}{8pt}\selectfont
    \begin{tabular}{ll|cc}
    \toprule
       $\theta_m$  & $\theta_n$ & $s(\texttt{eot}; \theta_m,\theta_n)$ &$s(r; \theta_m,\theta_n)$ \\ \midrule
       
        tiny.en & base.en & 12.13 & $\underset{\pm1.40}{3.50}$\\

        tiny.en & small.en & 13.29 & $\underset{\pm1.80}{4.13}$\\

        tiny.en & medium.en & 9.14 & $\underset{\pm1.37}{3.20}$\\ \midrule

        base.en & small.en & 19.72 & $\underset{\pm1.93}{5.61}$\\

        base.en & medium.en & 6.65 & $\underset{\pm1.42}{4.32}$\\ \midrule

        small.en & medium.en & 13.40 &$\underset{\pm1.78}{3.81}$\\ \midrule\midrule

        tiny & base & 6.54 & $\underset{\pm0.35}{1.24}$\\
        tiny & small & 4.57 &$\underset{\pm1.22}{5.46}$\\
        tiny & medium & 4.21 & $\underset{\pm2.04}{6.16}$\\ \midrule

        base & small & 9.67 & $\underset{\pm1.19}{5.51}$\\
        base & medium & 9.41 &$\underset{\pm2.12}{7.27}$\\ \midrule

        small & medium & 3.52 &$\underset{\pm1.57}{3.03}$\\
        \bottomrule
    \end{tabular}
    \caption{Measuring theoretical potential transferability of muting attacks between models.}
    \label{tab:theoretical-transfer}
\end{table}

Nevertheless, next we explore methods to learn a universal attack segment that is able to transfer across the different models: we explicitly train the attack audio segment by considering multiple models at the same time during the training of the attack segment. We also explore initializing the attack audio segment with the optimal audio segments for single target models. The results are presented in Table \ref{tab:emp-learn-multi}. As expected from the above analysis, it is clear that it is difficult to learn an attack that can transfer across multiple models. However, we are able to obtain an audio attack segment that can transfer between the tiny and base model (when training to attack tiny, base and small), or between the tiny and medium models. Overall, this section has demonstrated that analytically there is little potential of a muting attack that can transfer between models because there is no real sound for the acoustic realization of the $\texttt{<|endoftext|>}$ token, and therefore a specific acoustic realization is learnt for each specific target model.

\begin{table}[htb!]
    \centering
    \small
    \fontsize{8pt}{8pt}\selectfont
    \begin{tabular}{p{2cm}|l|l|cc}
        \toprule
        Trn models & Init & eval model & \multicolumn{2}{c}{Performance}  \\
       && & $\varnothing $ & $\texttt{asl}$ \\
        \midrule
        
        tiny.en& rand & tiny.en  &  99.7 & 0.06 \\
        & & base.en & 0.0 & 17.9  \\ \midrule
        
        tiny.en, base.en & rand& tiny.en & 99.42 & 0.160 \\
        & & base.en & 0.00 & 17.79 \\ \cmidrule{2-5}

        & base.en & tiny.en & 0.00 & 18.07\\
        && base.en  & 98.81 & 0.267\\ \midrule

        tiny.en, base.en, small.en& rand &  tiny.en & 98.43 &0.590\\
        &  & base.en & 99.12 & 0.227 \\
        & & small.en & 0.00 & 17.75\\ \cmidrule{2-5}

          & small.en&  tiny.en & 0.00 & 17.88\\
        &  & base.en & 0.00 & 17.79\\
        & & small.en & 99.22 & 0.070\\ \midrule

        tiny.en, base.en, small.en, medium.en & rand & tiny.en &  95.10 & 1.52\\
        & & base.en &  0.00 & 17.53\\
        & & small.en &  0.00 & 17.50\\
        & & medium.en & 98.33 & 0.670\\
        \bottomrule
    \end{tabular}
    \caption{Training universal muting attack on multiple models. Training epochs is the maximum number of epochs required for each of the \textit{train} (Trn) models when attacked individually. Initialization of the audio attack segment is either random or a previously targeted model. The best of 3 seeds is selected to obtain the most transferable attacks.}
    \label{tab:emp-learn-multi}
\end{table}

\section{Saliency Analysis Plots} \label{sec:app-sal-plots}

\begin{figure*}[htb!]
     \centering
     \begin{subfigure}[b]{0.45\textwidth}
         \centering
         \includegraphics[width=\textwidth]{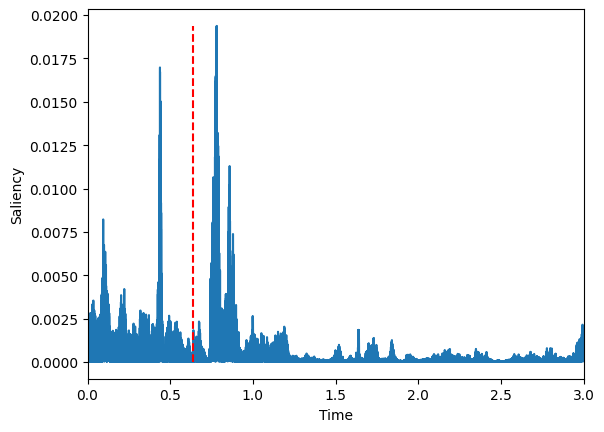}
         \caption{Successful Attack}
     \end{subfigure}
     \hfill
     \begin{subfigure}[b]{0.45\textwidth}
         \centering
         \includegraphics[width=\textwidth]{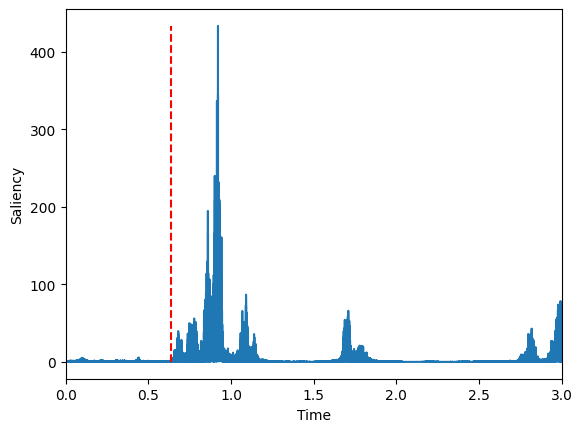}
         \caption{Unsuccessful Attack}
     \end{subfigure}
    
        \caption{Frame-level saliency plot, where the first 0.64-second represents the universal acoustic attack segment and the remainder is a randomly sampled speech signal (truncated to a total length of 3 seconds) for the target model Whisper medium.en was un/successfully muted by the universal adversarial attack.}
        \label{fig:app-sal-frame}
\end{figure*}

In the results in the main paper, we conduct a saliency analysis as per Section \ref{sec:saliency}, to better understand the mechanism of the adversarial attack for when it succeeds relative to when it fails. In Table \ref{tab:saliency} we report the average saliency for the adversarial segment, $\tilde s$ and the average saliency for the speech signal, $s$. It is also useful to visualize the frame-level saliency, to understand how the saliency changes from the adversarial segment per frame to the speech signal. In Figure \ref{fig:app-sal-frame} we have selected two random speech samples: one for which the universal acoustic attack succeeded, and one for which it failed. As we would expect, we observe two very different frame-level saliency patterns. For a successful attack, the saliency is heavily concentrated in the adversarial segment and then suddenly decays for the speech signal, whereas for the failed samples, the converse appears to be true.

\section{Spectrogram Plots}

\begin{figure*}[htb!]
     \centering
     \begin{subfigure}[b]{0.24\textwidth}
         \centering
         \includegraphics[width=\textwidth]{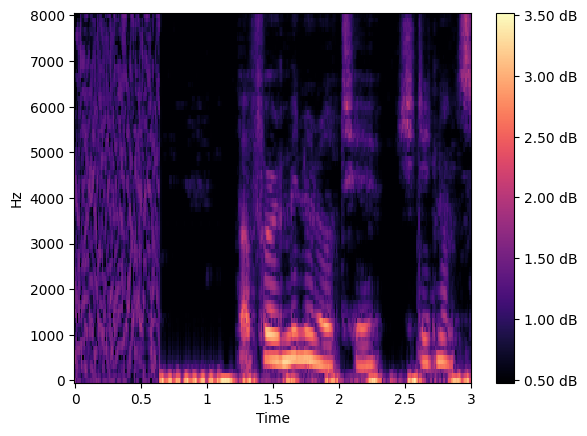}
         \caption{tiny.en}
     \end{subfigure}
     \hfill
     \begin{subfigure}[b]{0.24\textwidth}
         \centering
         \includegraphics[width=\textwidth]{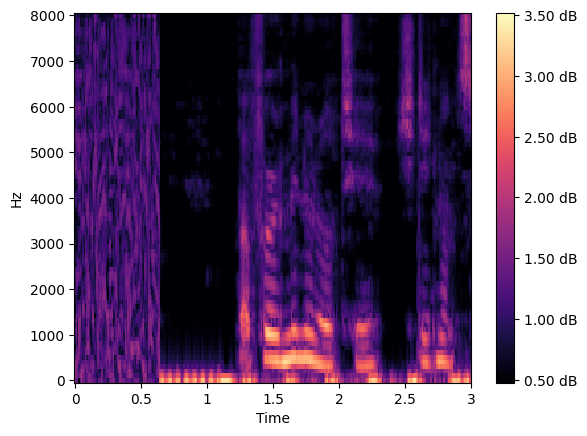}
         \caption{tiny}
     \end{subfigure}
     \hfill
     \begin{subfigure}[b]{0.24\textwidth}
         \centering
         \includegraphics[width=\textwidth]{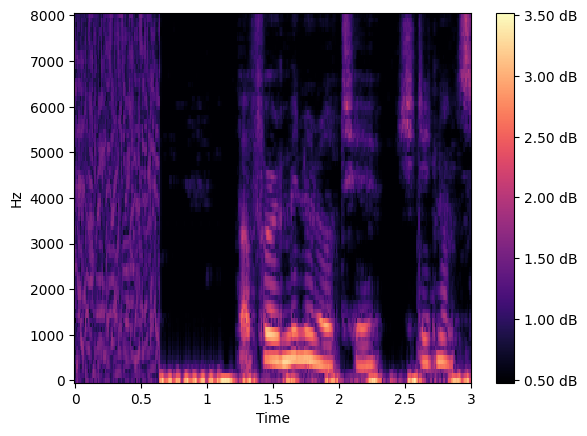}
         \caption{base.en}
     \end{subfigure}
     \hfill
     \begin{subfigure}[b]{0.24\textwidth}
         \centering
         \includegraphics[width=\textwidth]{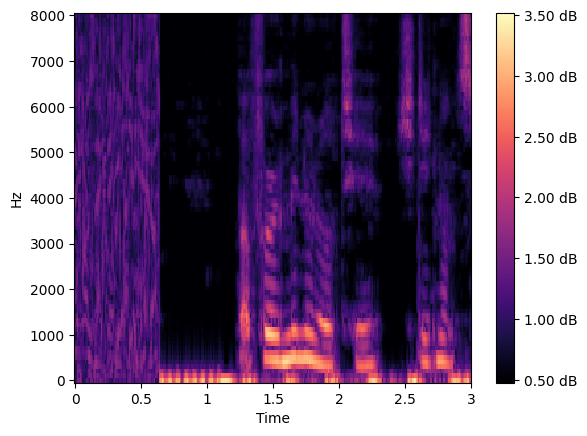}
         \caption{base}
     \end{subfigure}
     \hfill
          \begin{subfigure}[b]{0.24\textwidth}
         \centering
         \includegraphics[width=\textwidth]{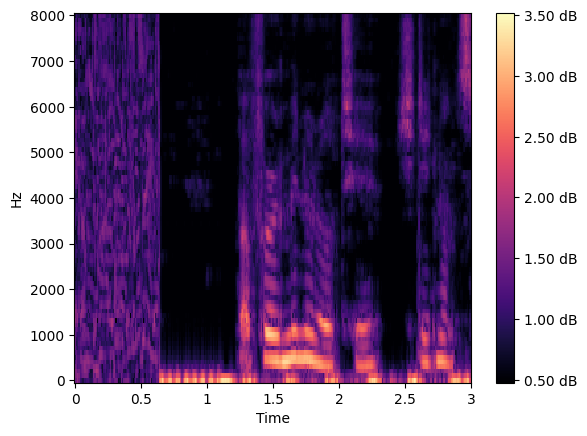}
         \caption{small.en}
     \end{subfigure}
     \hfill
     \begin{subfigure}[b]{0.24\textwidth}
         \centering
         \includegraphics[width=\textwidth]{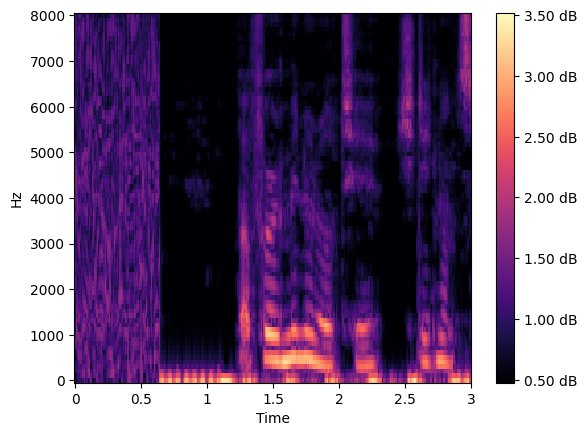}
         \caption{small}
     \end{subfigure}
     \hfill
     \begin{subfigure}[b]{0.24\textwidth}
         \centering
         \includegraphics[width=\textwidth]{latex/Figures/medium_spect_e0.02.png}
         \caption{medium.en}
     \end{subfigure}
     \hfill
     \begin{subfigure}[b]{0.24\textwidth}
         \centering
         \includegraphics[width=\textwidth]{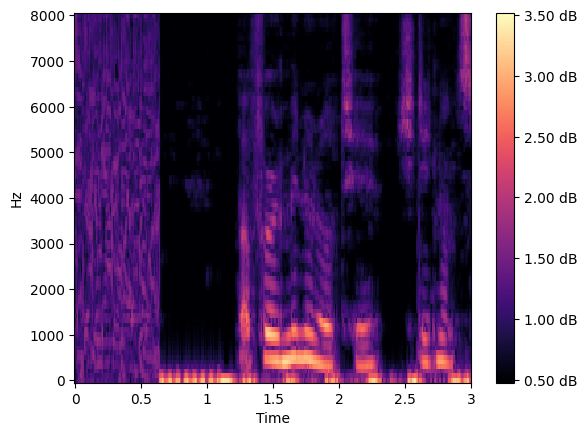}
         \caption{medium}
    \end{subfigure}
        \caption{Mel spectrogram of universal acoustic segment (0.64s) prepended to a random speech sample from LBS dataset (truncated to a total length of 3s) for different target Whisper models.}
        \label{fig:app-spectrograms}
\end{figure*}

\begin{figure*}[htb!]
     \centering
     \begin{subfigure}[b]{0.32\textwidth}
         \centering
         \includegraphics[width=\textwidth]{latex/Figures/tiny_spect_e0.02.png}
         \caption{$\epsilon=0.02$}
     \end{subfigure}
     \hfill
     \begin{subfigure}[b]{0.32\textwidth}
         \centering
         \includegraphics[width=\textwidth]{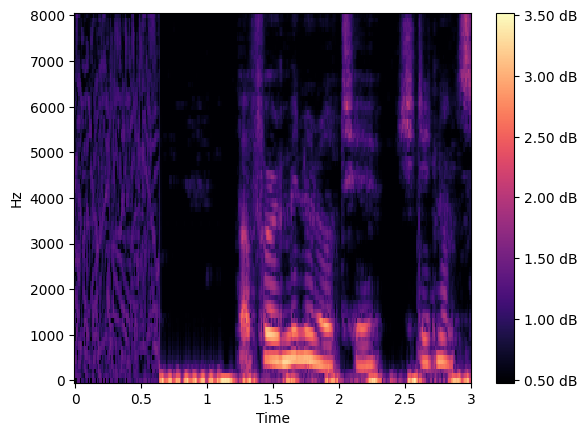}
         \caption{$\epsilon=0.01$}
     \end{subfigure}
     \hfill
     \begin{subfigure}[b]{0.32\textwidth}
         \centering
         \includegraphics[width=\textwidth]{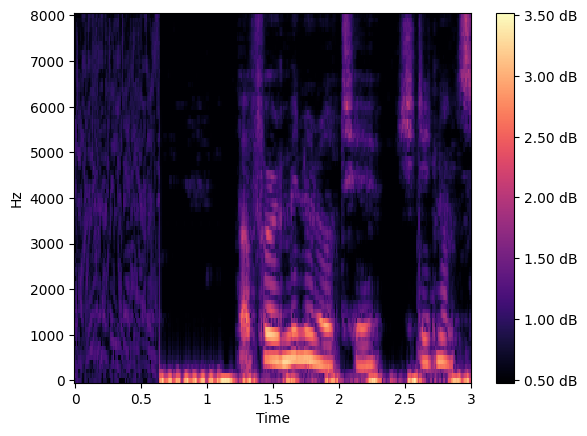}
         \caption{$\epsilon=0.005$}
     \end{subfigure}
    
        \caption{Mel spectrogram of universal acoustic segment (0.64s) prepended to a random speech sample from LBS dataset (truncated to a total length of 3s) for different amplitude constraints $\epsilon$ for the target model Whisper tiny.en.}
        \label{fig:app-spectrograms-e}
\end{figure*}

Log-mel spectrograms give a frequency-time representation of audio signals in a manner that can help to interpret the nature of the audio signal. The main paper gives an example of a log-mel spectrogram for an audio signal where a universal acoustic segment (learnt for the Whisper medium model) has been prepended to a specific speech signal. For reference, in this section we provide the remaining spectrograms. Figure \ref{fig:app-spectrograms} gives the spectrograms for the universal acoustic adversarial segments learnt for each target Whisper model, where the adversarial segment is of length 0.64-seconds and a maximum amplitude of $\epsilon=0.02$, to satisfy the imperceptibility constraint of Equation \ref{eqn:const}. Next, in Figure \ref{fig:app-spectrograms-e} we present the spectrograms for different universal adversarial attack segments with a different strictness of the amplitude constraint, $\epsilon$. As would be expected, the stricter the constraint the lower the relative power of the adversarial segment relative to the speech signal.

\end{document}